\documentclass[letterpaper, 10 pt, conference]{ieeeconf}
\usepackage{amsmath,amssymb,amsfonts}
\usepackage{bbm}
\usepackage{siunitx}
\usepackage{cite}
\usepackage{graphicx}
\usepackage{algorithm}
\usepackage[noend]{algpseudocode}
\usepackage{hyperref}
\usepackage{bbold}
\usepackage{xcolor}
\usepackage{caption}
\usepackage{subcaption}
\usepackage{bm}
\usepackage[belowskip=-9pt,aboveskip=3pt]{caption}
\definecolor{green}{RGB}{80,190,110}
\definecolor{purple}{RGB}{220,60,220}

\IEEEoverridecommandlockouts                              

\title{\LARGE \bf
The Grasp Loop Signature: A Topological Representation for Manipulation Planning with Ropes and Cables}

\author{Peter Mitrano$^{1}$ and Dmitry Berenson$^{1}$
\thanks{This work was supported in part by Toyota Research Institute, the Office of Naval Research Grant N00014-21-1-2118, and NSF grants IIS-1750489, IIS-2113401, and IIS-2220876.$^{1}$Department of Robotics, University of Michigan}%
}

\newcommand{\sig}{{\mathcal{G}_L}}
\newcommand{\signature}{$\sig$-signature}
\newcommand{\state}{s}
\newcommand{\q}{q}
\newcommand{\loc}{l}
\newcommand{\locs}{\bm{l}}
\newcommand{\kp}{\loc_k}
\newcommand{\goal}[1]{#1_\text{goal}}
\newcommand{\goalPoint}{\goal{p}}
\newcommand{\goalRadius}{\goal{d}}
\newcommand{\goalSig}{\goal{\sig{}}}
\newcommand{\qvel}{\dot{q}}

\newcommand{\skels}{\textbf{S}}
\newcommand{\skel}{\mathit{S}}
\newcommand{\sk}{\textbf{s}}

\newcommand{\maxIters}{i^\text{max}}

\newcommand{\planGrasp}{\textit{PlanGrasp}}
\newcommand{\indic}{\mathbb{1}}
\newcommand{\isGrasping}{\bm{\indic_{g}}}
\newcommand{\isGraspingI}{\indic_{\bm{g},i}}
\newcommand{\ncon}{n_\text{con}}
\newcommand{\blist}{\mathcal{B}}

\newcommand{\rev}[1]{#1}

\begin{document}

\maketitle
\thispagestyle{empty}
\pagestyle{empty}

\begin{abstract}

This paper studies robotic manipulation of deformable, one-dimensional objects (DOOs) like ropes or cables, which has important potential applications in manufacturing, agriculture, and surgery. In such environments, the task may involve threading through or avoiding becoming tangled with other objects. Grasping with multiple grippers can create closed loops between the robot and DOO, and if an obstacle lies within this loop, it may be impossible to reach the goal. However, prior work has only considered the topology of the DOO in isolation, ignoring the arms that are manipulating it. Searching over possible grasps to accomplish the task without considering such topological information is very inefficient, as many grasps will not lead to progress on the task due to topological constraints. Therefore, we propose the \signature{} which categorizes the topology of these grasp loops and show how it can be used to guide planning. We perform experiments in simulation on two DOO manipulation tasks to show that using the \signature{} is faster and more successful than methods that rely on local geometry or \rev{additional} finite-horizon planning. Finally, we demonstrate using the \signature{} in a real-world dual-arm cable manipulation task. 

\end{abstract}

\section{Introduction}

Manipulation planning for deformable one-dimensional objects (DOOs) like ropes and cables is challenging due to the high-dimensional state representation of these objects and the cost of simulating their motion. Furthermore, most tasks benefit from multiple arms to control DOO shape and avoid becoming tangled with the environment. Therefore, the planner needs to consider the DOO, the arms manipulating it, and the environment. A task and motion planning (TAMP) approach to this problem would decompose planning into a grasp selection problem and a motion planning problem for the DOO given a specific grasp, as in \cite{RitaCableRouting,NairRope,YanContrastiveRope}. However, the DOO planning problems are often expensive to solve. To reduce the space of grasps we need to search, we borrow the idea of a \textit{signature} from the field of topology.

\begin{figure}
    \centering
    \includegraphics[width=0.75\linewidth]{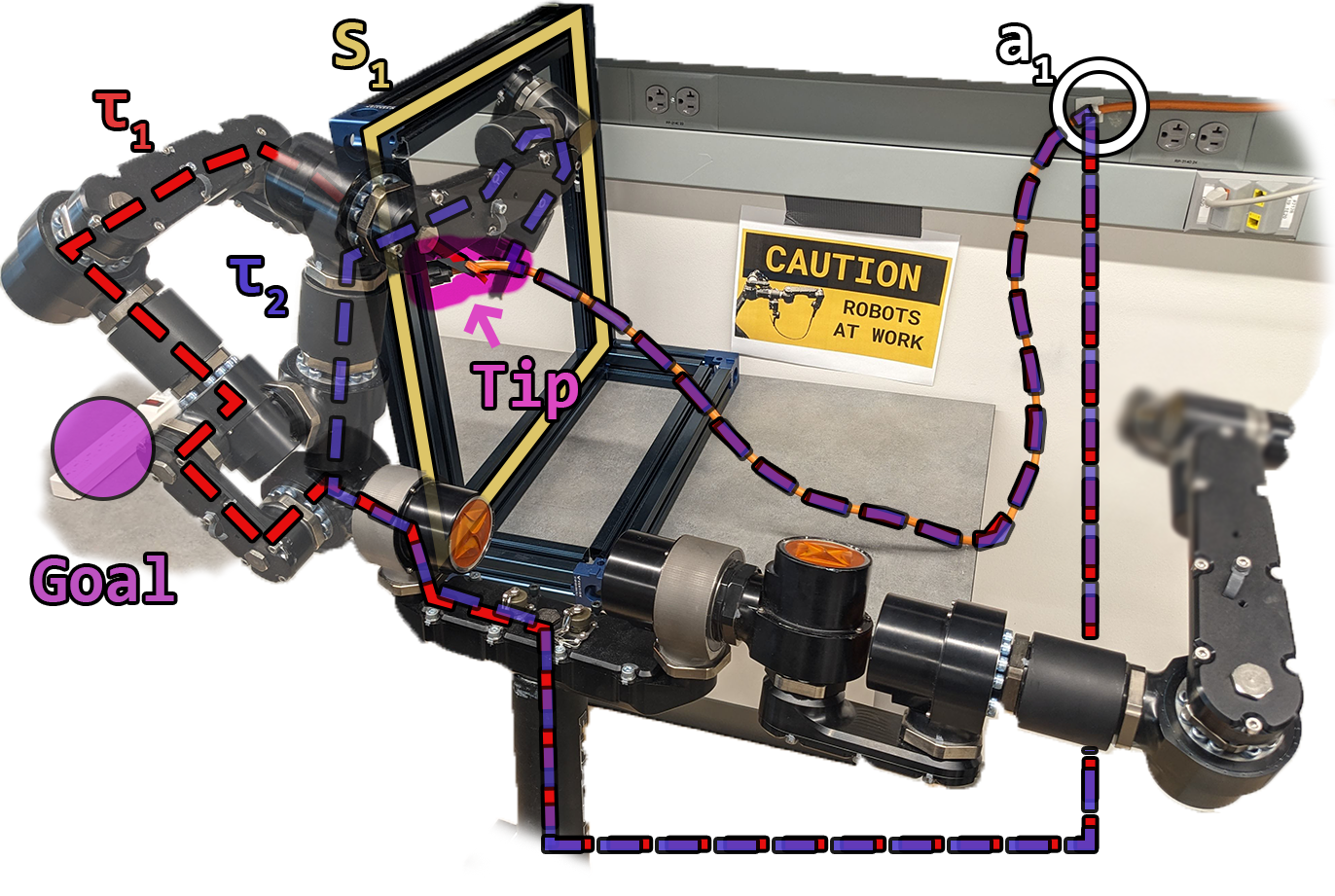}
    \caption{Annotated image of our real world cable threading setup. The red dashed line shows a grasp loop $\tau_1$ that is linked with the skeleton $\skel_1$. The blue grasp loop is not linked with $\skel_1$. This distinction is captured by the proposed \signature{} and is used in planning.}
    \label{fig:titleFig}
\end{figure}

To explain what this signature represents, consider how the robot should grasp the tip of the cable in Figure \ref{fig:titleFig}. By grasping we form a loop, which we call a \textit{grasp loop} and show as blue and red dashed lines in Figure \ref{fig:titleFig}. It is possible to grasp either around the left side or the right side of the frame, but these two grasps are categorically different in that we cannot smoothly deform from one to the other without breaking the grasp or the frame. The frame also forms a loop, called an obstacle loop \rev{($\skel_1$)}. When grasping from the left (red), these two loops are linked, but when grasping from the right (blue) they are not. Our key insight is that the robot, DOO, and environment form a \textit{graph of grasp loops} and we can use this graph to construct a signature, \signature{}, which captures topological information relevant for planning. To be clear, we do not address knots in the DOO. Our work is complimentary to work on tying or untying knots \cite{WakamatsuKnots2005, Saha07, UntanglingFull, WeifuKnots}.

The main contribution of this paper is the \signature{} which compactly represents the \rev{topological relationship between the object, the arms manipulating it, and the environment.} We \rev{demonstrate that this signature is useful for manipulation planning with DOOs}. In simulation, we show that methods using the \signature{} outperform baselines and ablations which search for grasps without using topological information. Finally, we demonstrate a threading and point reaching task on a physical robot. Videos and animations can be found on our Project Website\footnote{\href{https://sites.google.com/view/doo-manipulation-signature/home}{https://sites.google.com/view/doo-manipulation-signature/home}}.

\section{Related Work}

\textit{Topology in Motion Planning:}
Topology and homotopy have been used in path planning for flying and driving robots \cite{Bhattacharya11,Bhattacharya12}, as well as tethered robots \cite{TetherHomotopy}. \cite{TetherHomotopy} operates only in 2D and \cite{PDR_Jaillet} considers an approximation of homotopy for 3D path planning. \cite{Bhattacharya11} introduces a simple-to-compute and exact signature for characterizing the homotopy of 3D paths with respect to 3D obstacles with holes in them, called the h-signature. We build on \cite{Bhattacharya11} to define our \signature{}.

\textit{Manipulation Planning for Deformable Objects:}
Prior work on knot tying and untying has also applied knot theory to DOO manipulation \cite{WakamatsuKnots2005, Wakamatsu2006Untangling, Saha07, WeifuKnots, UntanglingHulk, UntanglingDescriptors, UntanglingFull}. These methods use planar crossing representations, which project a curve into a specified plane and count the sequence and type of crossings. \cite{Saha07} used this method for robotic knot tying, and extended this to tying around obstacles by specifying connections between obstacles and the DOO. However, these methods do not consider how the manipulator effects the topology, and fail in some cases with non-planar obstacles. A method for threading surgical needles was proposed in \cite{Weifu}, but uses floating grippers and does not address planning for the robots' arms or obstacles, and is limited to tight-tolerance insertion tasks. \cite{TetheredToolManipulation, UnreliableMitrano2021, UnreliableDale2019, DataAugmentation2022, FOCUS2023} all address manipulation planning for DOOs assuming the grasp is fixed, which is complementary to this work.

\textit{Grasping and Regrasping:}
With rigid grasping, the challenge is primarily in achieving a stable grasp \cite{DexNet1,DexNet2}. With deformables, the challenge is where to grasp, as studied in cloth smoothing or folding \cite{ClothSmoothingHoque,ClothSmoothingLin,ClothSmoothingWu}. These works use pick-and-place primitives with a single manipulator, which is too restrictive for many tasks. In contrast, we use a dual-arm manipulator and use joint velocities as our action space. In \cite{Zhang2022}, a dual arm manipulator autonomously dresses a mannequin. Their method for grasp planning is based on learned visual models of the garment, and only considers grasps near keypoints such as the elbow or shoulder. The methods in \cite{RitaCableRouting,NairRope,YanContrastiveRope} plan grasps on the DOO, but they assume the rope is planar (flat on a table), use one manipulator, and do not consider obstacles for the manipulator. \cite{Simeon04} describes a method that produces pick, place, and sliding paths in the configuration space without explicit task planning. However, this method does not address underactuated kinodynamic systems such as DOOs.

\section[Defining the signature]{Defining the \signature{}}
\label{sec:defHomotopy}

\begin{figure}
    \centering
    \includegraphics[width=\linewidth]{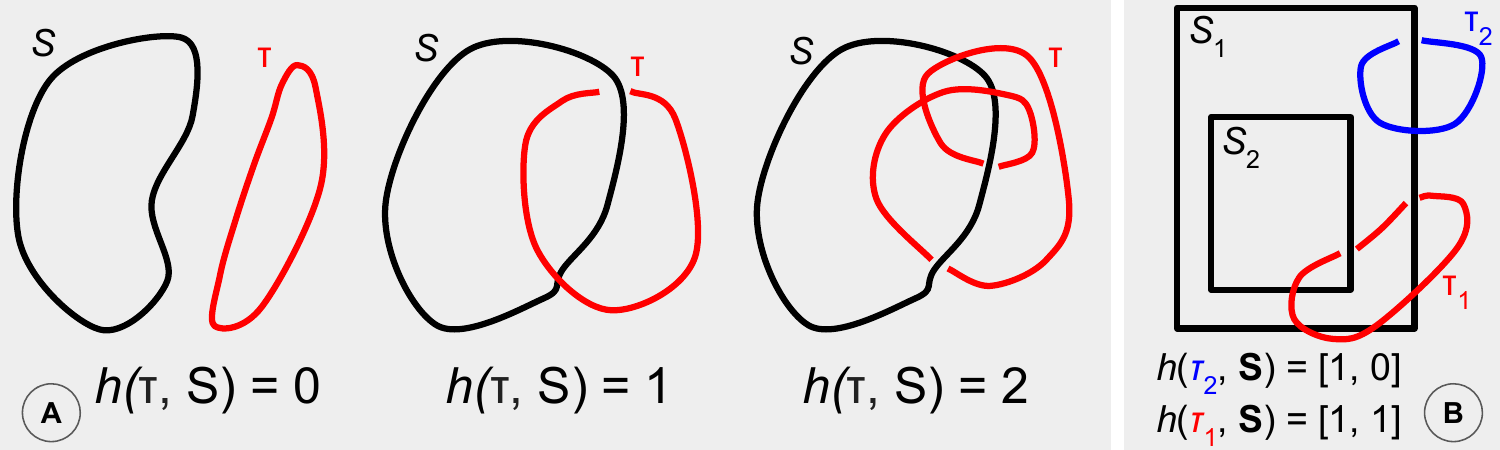}
    \caption{(A) Illustration of the h-signature for a loop representing the robot and DOO (red/blue) and a loop representing an obstacle (black). (B) Two examples of the h-signature for a skeleton with two obstacle loops $S_1$ and $S_2$.}
    \label{fig:simple_h}
\end{figure}

\subsection{Preliminaries}

We primarily use notation that is consistent with \cite{Bhattacharya11}. We call a closed one-dimensional curve in 3D a \textit{loop}. The environment is assumed to be decomposed into a \textit{skeleton} made up of multiple \textit{obstacle loops} $\skels = \{\skel_1,\dots,\skel_n\}$. Each obstacle loop is made up of line segments $\skel_i=\{\textbf{s}_i^1,\dots,\textbf{s}_i^{n_i}\}$. In practice, the skeleton can either be specified manually or computed automatically from a medial axis transform of a mesh or pointcloud of the environment. \rev{The state $\state=(q,o)$ contains the robot state $q$ (joint angles) and the DOO state $o$ (ordered list of points in $\mathbb{R}^3$)}.

\cite{Bhattacharya11} plans paths that are in a given homotopy class or avoid a certain homotopy class. They compare two paths by considering the homotopy class of the closed loop $\tau$ formed by joining the two paths at their shared start and end points. For a path loop $\tau$ and an obstacle loop $\skel$, \rev{they} define the h-signature $h(\tau, \skel) \in \mathbb{Z}$, which counts the number of times $\tau$ passes through $\skel$. The sign of $h$ in this case is determined by the direction of $\tau$. The h-signature can be extended to a list of the h-signatures with respect to each obstacle loop in the skeleton $h(\tau, \skels)) = [h(\tau,\skel_1), \dots,h(\tau,\skel_n)]$. These cases are illustrated in Figure \ref{fig:simple_h}. The equation for computing $h(\tau,\skel)$ is reproduced from \cite{Bhattacharya11}. The point $\sk_i^{j'}$ is the point that follows $\sk_i^j$, and $r$ is a point on the loop $\tau$. The integration over $\tau$ is done numerically.

\begin{equation}
\label{eq:hsig}
\begin{split}
    h(\tau,\skel) = \frac{1}{4\pi} \int_{\tau}\sum_{j=1}^{n_i}\Phi(\sk_i^j,\sk_i^{j'}, r) \Delta r \\
    \Phi(\sk_i^j,\sk_i^{j'}, r) = \frac{1}{||d||^2} \Big( \frac{d\times p'}{||p'||} - \frac{d \times p}{||p||} \Big) \\
    p=\sk_i^j-r,\quad p'=\sk_i^{j'},\quad d=\frac{(\sk_i^{j'}-\sk_i^j)\times(p\times p')}{||\sk_i^{j'} - \sk_i^j||^2}
\end{split}
\end{equation}

We take this idea but apply it to grasp loops, instead of paths. Unlike in path planning, where the direction of $\tau$ matters, we only care how or whether loops are linked. Accordingly, we assert that $h$ is always non-negative.

\subsection[Computing the signature]{Computing the \signature{}}
\label{sec:defGL}

\begin{figure}
    \centering
    \includegraphics[width=\linewidth]{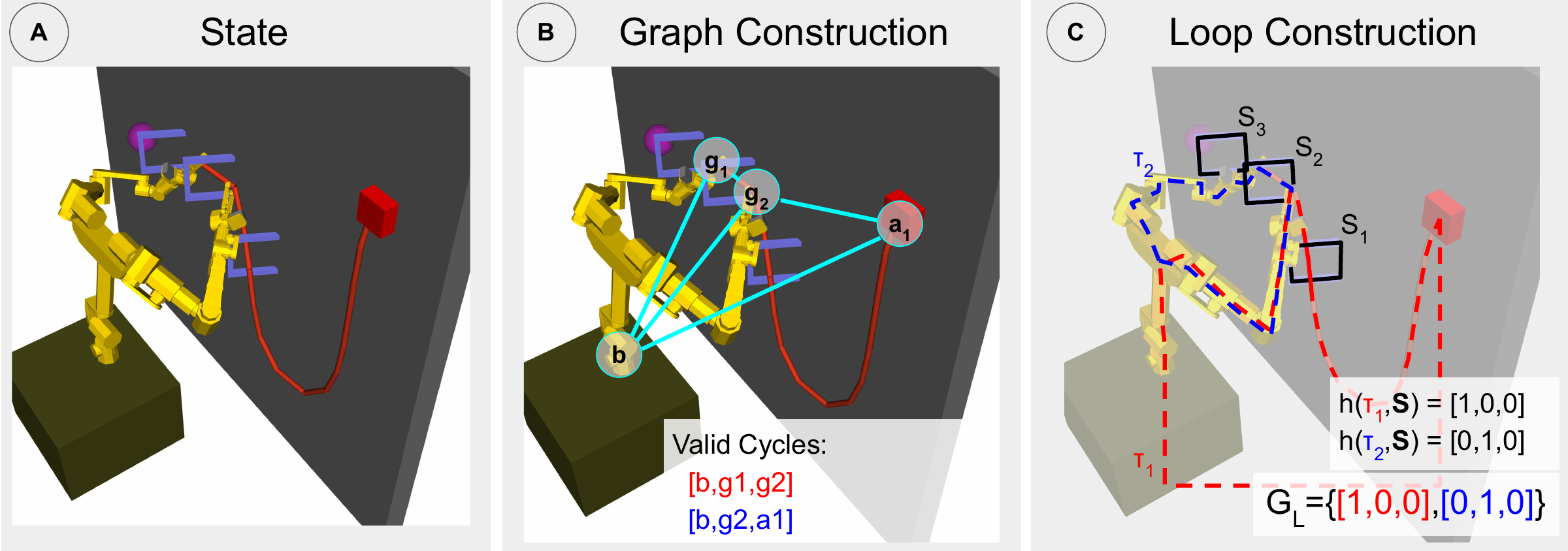}
    \caption{The process of constructing the \signature{}. (C) There are 2 grasp loops and 3 object loops, so the \signature{} is a set with two elements, and each element is a vector of 3 non-negative integers.}
    \label{fig:constructingSignature}
\end{figure}

\begin{figure}
    \centering
    \includegraphics[width=\linewidth]{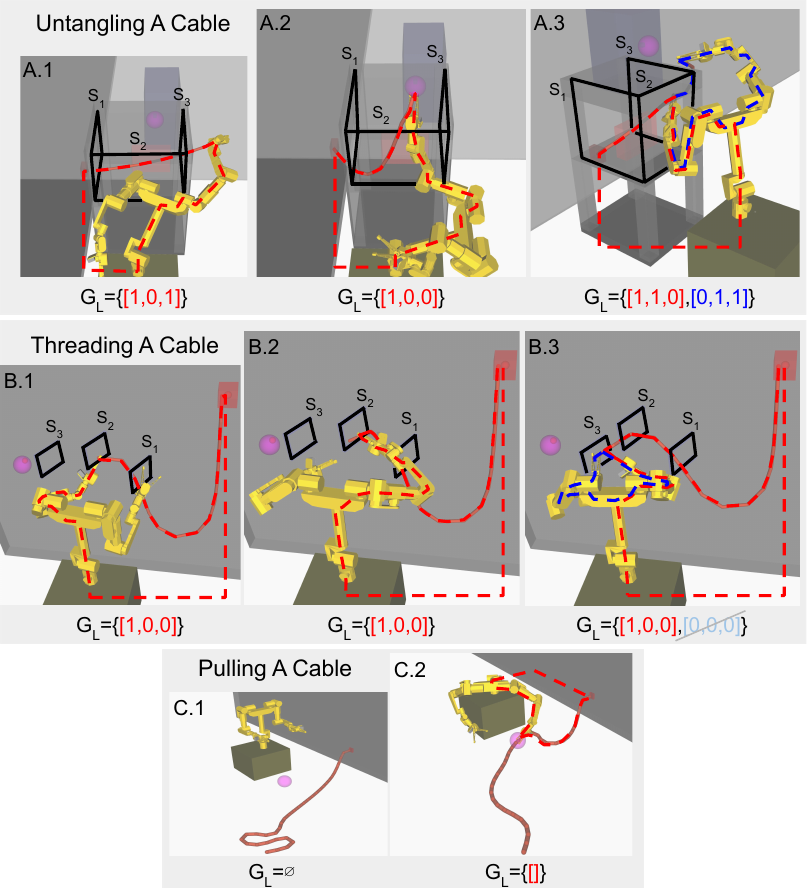}
    \caption{Example scenes and their $\sig$ values. \rev{The dashed blue and red lines are grasp loops, and the solid black lines are obstacle loops. (B.3) The blue grasp loop is omitted from the signature, as described in Section \ref{sec:defGL}. The purple sphere shows the goal region.}}
    \label{fig:examples}
\end{figure}

The \signature{} is composed of the h-signatures $h(\tau, \skels)$ of grasp loops $\tau$ formed by the robot and  DOO.
The grasp loops are constructed based on a graphical model of the state where vertices are the robot base, its grippers, and attach points, and edges are paths between them. Attach points are used to represent locations on the DOO which are fixed relative to the robot (e.g. plugged into the wall or \rev{tethered to} the robot itself). Figure \ref{fig:constructingSignature} illustrates how the graph construction step works. The robot base vertex is connected to all gripper and attach vertices because it connects to the grippers directly (via the robot geometry) and the attach points indirectly (via the environment). Edges connect grippers/attach points to one another if they are adjacent on the DOO. In Figure \ref{fig:constructingSignature}, the vertices $(g_1,g_2)$ are adjacent, as are $(g_2,a_1)$, but $(g_1,a_1)$ are not.

From \rev{this graph}, we extract all cycles of 3 vertices which contain a gripper, and convert each cycle to a grasp loop $\tau$. \rev{We consider only length-3 cycles containing a gripper in our signature because these contain the relevant information for planning how to manipulate the DOO. Including longer cycles would add redundant information. Likewise, cycles not containing a gripper are omitted for compactness, since we assume that attach points cannot be changed by the planner.} To make a grasp loop from a cycle, we concatenate the 3D paths represented by the cycles' edges. This requires a skeletonized representation of the robot geometry, which can be constructed from the kinematic tree and the origins of the links, as well as the points representing the DOO. A path between the robot base and an attach point ($(b,a_1)$ in Figure \ref{fig:constructingSignature}) can be chosen arbitrarily, as long as it is the same for all states.

For each grasp loop \rev{$\tau_i$}, we compute the associated h-signature $h(\tau_i, \skels)$. The \signature{} of the state, denoted $\sig(\state)$, is the \textit{multiset} of the h-signatures of each grasp loop. In a multiset the order does not matter, but elements may repeat. The number of repetitions of an element is called its multiplicity. Two multisets are equivalent if their elements and multiplicities are equal. Preserving repetitions in the \signature{} allows us to represent multiple grasp loops that go through the same obstacle loop.

This may result in a grasp loop \rev{$\tau$} containing two grippers \rev{for which} $h(\tau,\skels)=\bm{0}$ (i.e. not linked $\skels$). The \rev{blue dashed grasp loop shown in Figure \ref{fig:examples} B.3 shows this}. Releasing one of the grippers does not categorically change what we can do with the object, and neither would grasping with an additional gripper right next to two already grasping. Therefore, if there is a cycle with $h(\tau,\skels)=\bm{0}$ containing two grippers, one of the grippers is removed from the graph and the process restarts from the cycle extraction step.

\subsection{Computational Complexity}

The complexity of computing the \signature{} can be written in Big-O notation based on the number of skeletons $n_s$, line segments in the skeleton $l_s$, arms and/or attach points $n_a$, and the length of the arms and/or DOO $l_a$. \rev{Due to the rules of construction defined above, in the worst case where all arms are grasping, the graph of grasps loops is a fan graph ($F_{1,n_a}$) \cite{FanGraph}. In the base case of $n_a=2$, the graph has 3 vertices and one cycle of length 3. Adding another vertex adds one cycle, so in the worst case $n_a$ arms/attach points create $n_a-1$ loops.} Each cycle (loop) is compared with each skeleton, and the number of comparisons scales linearly with both the number of line segments and the length of the loop, giving a total complexity of $O\big((n_a-1) n_s l_s l_a\big)$. Our Python implementation, using the NetworkX library \cite{NetworkX} for computing the \signature{} for a state, takes $\leq$10ms in the Untangling and Threading environments.

\section[DOO Manipulation with the signature]{DOO Manipulation with \signature{}}

\subsection{Problem Statement}

\newcommand{\goalFunc}{\goal{\indic}}
\newcommand{\goalCost}{\goal{C}}
\newcommand{\graspCost}{C_\text{grasp}}

In this section, we define the DOO manipulation problem which our proposed planning method addresses. In our experiments, the robot has two 7-dof arms attached to a 2-dof torso with parallel-jaw grippers, but the \signature{} can be applied to other robot morphologies. We assume we have a complete geometric model and skeleton of the environment. When manipulating with the current grasp, the action space is joint velocities $\qvel$. We describe points on the DOO primarily by their location $\loc\in[0,1]$, where $\loc=0$ is one end of the DOO and $\loc=1$ is the other. Each location also corresponds to a point $p(\loc)\in\mathbb{R}^3$. Grasps are represented by a vector of locations $\locs=[\loc_1,\loc_2]$, one for each gripper. A set of grasp locations $\locs$ must also be paired with a collision-free motion of the robot to the new grasp locations, which may be reachable by many distinct joint configurations.

The goal of the manipulation is to bring a \textit{keypoint} $\kp$ on the DOO to a goal region with position $\goalPoint$ and radius $\goalRadius$. This is a useful skill for plugging in cables, or for using tools with an attached cable or hose, and more complex tasks like cable harnessing can be described as a sequence of these point reaching goals. Additionally, one can specify a desired \signature{} for the goal $\goalSig$. This type of DOO manipulation is complementary to tying or untying knots, which has been addressed in prior work \cite{Saha07,UntanglingFull,WeifuKnots}.

\subsection{DOO Point Reaching Method}

Algorithm \ref{alg:pointReaching} describes our method for point reaching tasks. 
This algorithm serves to demonstrate the utility of the \signature{} in planning. It uses cost functions that are designed for our robot and our scenarios, and these functions may need to be modified for other applications. Given the current grasp, we use MPPI \cite{mppi} to find an action $\qvel$ that minimizes the goal cost $\goalCost$, shown in Eq.~\eqref{eq:pointReachingCost}. MPPI runs until the goal is reached or progress stops. If progress stops, we plan and execute a grasp change, and resume running MPPI. This process is repeated until the goal is reached (trial success) or for $\maxIters$ iterations (trial failure). For both MPPI and grasp planning, we model the dynamics of the robot, rope, and obstacles in MuJoCo \cite{mujoco}. 

\begin{algorithm}[t]
\caption{DOO Point Reaching with the \signature{}}
\label{alg:pointReaching}
\begin{algorithmic}[1]
\Procedure{pointReaching}{$\state,n_x,\kp,\goalPoint,\goalRadius$}
\For{$i < \maxIters$}
    \State $\qvel = $MPPI$(\state,\goalPoint, \goalCost)$
    \State $\state = f(\qvel)$  \Comment{Execute and get state}
    \If{$||\goalPoint - p(\kp)|| < \goalRadius$}
        \State break
    \EndIf{}
    \If{trapped} 
        \State $d_0 = \min(|\locs_0-\kp|) $ \Comment{Initial geodesic}
        \State $\locs^* = \planGrasp(\state, n_x, \graspCost, \kp, \rev{\blist)}$
        \State $d^* = \min(|\locs^*-\kp|) $
        \If{$d^* \geq d_0$} \Comment{Unable to grasp closer}
            \State Add $\sig(\state)$ to $\blist$
            \State $\locs^* = \planGrasp(\state, n_x, \rev{\graspCost, \kp, \blist)}$
        \EndIf{}
        \State ExecuteGraspChange$(\locs^*)$
    \EndIf{}
\EndFor
\EndProcedure
\end{algorithmic}
\end{algorithm}

The method for determining if MPPI is trapped, called trap detection, is adapted from \cite{TAMPC}. Trap detection operates on a window of recent joint configurations $q_1,\dots,\q_T$, and computes the average one-step state difference $\bar{q} = \frac{q_T-q_1}{T}$ and keeps a running maximum of this value $\bar{q}^+$ over the trial. MPPI is considered trapped when the ratio $\frac{\bar{q}}{\bar{q}^+}$ is below a threshold (0.2-0.3 in our experiments).

The goal cost used for MPPI is shown in Equation \eqref{eq:pointReachingCost}, where the state $\state$ is used to compute the grasp locations $\locs$, grasping state $\isGrasping$, keypoint position $p(\kp)$, grasp positions $p(\locs)$, and number of contacts $\ncon$.

\begin{equation}
    \label{eq:pointReachingCost}
    \begin{split}
        \goalCost(\state,\qvel) = || p(\kp) - \goalPoint || + 
        \rev{\alpha_1 \sum_{i=1}^{n_a}\isGraspingI || p(\loc_i) - \goalPoint ||} + \\
        \alpha_2\sqrt{\ncon} + \alpha_3 || \qvel ||
    \end{split}
\end{equation}

The first term in Eq.~\ref{eq:pointReachingCost} brings the keypoint $p(l_k)$ towards the goal $\goalPoint$. The second term provides a reward for moving \rev{any grippers which are grasping towards the goal. $n_a$ is the number of arms, $\isGrasping$ is a binary vector ($\isGraspingI \in \isGrasping$) indicating which grippers are grasping, and $\loc_i\in\locs$ are the current grasp locations.} This term is useful when the DOO is slack and the keypoint cannot be pulled directly (See Figure \ref{fig:examples} C). The third term penalizes collision between the robot and environment, based on the number of contacts $\ncon$ reported by the dynamics. Finally, the fourth term penalizes high joint velocities to encourage smooth motion. The hyperparameters $\alpha_{1,2,3}$ were selected to prioritize collision avoidance first, then bringing the keypoint to the goal.

In \planGrasp{} we sample $n_x$ grasps ($\approx$50) and choose the best one. Grasps are \rev{sampled by first} choosing a strategy for each gripper. The possible strategies are \texttt{STAY}, \texttt{GRASP}, \texttt{MOVE}, or \texttt{RELEASE}. \rev{\texttt{STAY} means the gripper does not change its grasp, or remains not grasping. \texttt{GRASP} means the gripper is not grasping and should create a new grasp. \texttt{MOVE} means the gripper is currently grasping, but should move to a new grasp location. \texttt{RELEASE} means the gripper is currently grasping and should release.} For the \texttt{GRASP} or \texttt{MOVE} strategies, we sample a location $\loc \in [0,1]$. At least one gripper must be grasping. For each candidate grasp, we simulate release and grasp dynamics using MuJoCo. Modeling grasping using friction and caging is challenging, so we instead use equality constraints between the rope and the grippers that are grasping. MoveIt \cite{MoveIt} is used to find collision-free paths to move the grippers to desired grasp locations. The result is a candidate state $\state$ and collision free trajectory for each candidate grasp. We choose the grasp with the lowest cost according to Eq.~\ref{eq:graspCost}. With abuse of notation, we say the candidate state $\state$, change in state $\Delta \state$, and grasp state $\isGrasping$ are derived from the candidate grasp locations $\locs$.
            
\begin{equation}
    \label{eq:graspCost}
    \begin{split}
    \graspCost(\locs)= \indic{}_\text{feasible} + \indic_{\blist}(\state) + \indic_{\sig}(\state, \goalSig) + \\
    \isGrasping \cdot |\locs - \kp | + \beta_1 \Delta \state
   \end{split}
\end{equation}

The first term in Eq \ref{eq:graspCost} assigns a large penalty (e.g. 100) if no collision-free path to the grasp was found. The next two terms assign a large penalty based on the \signature{} of candidate state, either for matching a blocklisted signature \rev{(second term)} or for not matching the goal signature \rev{(third term)}. The fourth term encourages grasping near the keypoint, based on the geodesic distance for any grasping grippers. The final term penalizes the change in robot and DOO state. This results in shorter and faster grasps and is weighted by $\beta_1$ to be the least important term. The large penalties dominate the keypoint and state-change terms.

We use a blocklist of \signature{}'s to avoid retrying topological configurations in which we have failed to reach the goal.
\rev{Specifically, we blocklist the current \signature{} if the planner cannot find a grasp with lower geodesic cost (4th term in Eq \eqref{eq:graspCost}) than the current grasp (Alg \ref{alg:pointReaching} lines 8-12). In other words, we do not blocklist if it is possible to grasp closer to the keypoint. This avoids blocklisting the current \signature{} in the case that progress halts, not because of topological constraints, but because the grasp is too far away from the keypoint to control it. This is inspired by the idea of diminishing rigidity \cite{DiminishingRigidity}, which says that the control over a point on a deformable object decreases as the geodesic distance to the gripper increases.
In the case of multiple grippers, the initial grasp locations $\locs_0$ or new grasp locations $\locs^*$ may be a list of locations, in which case we use the $\min$ when computing the geodesic distance.}

\section{Applications}

We now describe how the above framework can be applied or adapted to DOO manipulation in three different environments, Pulling, Untangling, and Threading. \rev{We use a horizon of $H=15$ in MPPI.}

\subsection{Pulling Environment}

The Pulling environment contains a large hose attached to a wall. The scene is depicted in Figure \ref{fig:examples} C. The robot is initially not grasping the hose, and the head of the hose is out of the robot's reach. The goal region, shown as a purple sphere, is near the base of the robot on the floor. This environment requires regrasping to bring the \rev{head of the hose} to the goal, and demonstrates the behavior of the general method in the case where there are no skeletons, and no changes in the \signature{}.

When applying Alg \ref{alg:pointReaching} in this environment, the robot initially chooses a grasp as far down the DOO as it can reach, due to the geodesic cost term in Equation \ref{eq:graspCost}. Then, the gripper pulls towards the goal due to the second term in the MPPI cost \eqref{eq:pointReachingCost}. This brings more of the DOO within reach. When the gripper reaches the goal, the cost cannot be decreased and the controller slows to a stop. At this point, trap detection triggers regrasp planning. Since the DOO is now closer, a plan is found that reaches closer to the tip ($\kp=1$) than before. Because the grasp is closer to the tip, the current \signature{} is not blocklisted. This repeats until the grasp is close enough to the tip that it can be brought to the goal region. In the Pulling environment, our method succeeded in 25/25 trials, where each trial differs in the initial DOO configuration and the random seed used for sampling in planning.

\subsection{Untangling Environment}

\begin{table}
    \centering
    \begin{tabular}{ccccc}
        Method & Success & Wall Time (m) & Sim Time (m) \\ \hline
        \signature{} (ours) & \textbf{22/25} & \textbf{12 (5)} & 1.4 (1.1) \\
        Always Blocklist & \textbf{22/25} & 14 (7) & \textbf{1.3 (1.0)} \\
        No \signature{} & 10/25 & 20 (10) & 2.0 (1.3) \\
        TAMP50 & 15/25 & 142 (116) & 2.4 (1.7) \\
        TAMP5 & 9/25 & 34 (22) & 1.8 (1.2) \\
    \end{tabular}
    \caption{Results in the Untangle environment. Times in minutes are for the completion of the task, where Sim Time does not include planning time. Standard deviations are given in parentheses.}
    \label{tab:point_reaching}
\end{table}

The Untangle environment resembles a computer rack with a cable that needs to be plugged in. The scene is depicted in Figure \ref{fig:examples} A. One end of the DOO is fixed to the environment (e.g. plugged in elsewhere), and the robot is initially grasping some other location on the DOO. The robot often needs to regrasp several times in order to reach the goal. Unlike in the Pulling environment, the \signature{} can take on many different values depending on the configuration of the DOO and the grasp configuration. This demonstrates the utility of the \signature{} in planning when there is no goal \signature{}.

We evaluate Alg \ref{alg:pointReaching} on this task, and compare to an ablation that omits the two terms using the \signature{} from Eq \ref{eq:graspCost}. We call this method \textit{No \signature{}}. This often results in greedy re-grasping of the keypoint. We also evaluate a version called \textit{Always Blocklist}, where we blocklist the current \signature{} every time a trap is detected. Finally, we compare our proposed method to a method inspired by task and motion planning (TAMP), where $H$ additional steps of MPPI are simulated for each candidate grasp during planning and the final goal cost is used in place of cost terms relying on the \signature{}. We test two versions of this method with $H=5$ and $H=50$. Success rates and trial times are shown in Table \ref{tab:point_reaching}. Trials vary in the initial configuration of the robot, grasp location, DOO configuration, in the size of the computer rack, and in the location of the goal.

Methods using the \signature{} have the highest success rates and are faster than alternatives. \textit{Always Blocklist} has an equivalent success rate as the full proposed method, but prematurely abandons grasps that would lead to reaching the goal. Our method and the \textit{Always Blocklist} method each failed in 3 trials by trying too many unsuccessful grasps before $\maxIters$ was reached. The \textit{No \signature{}} ablation and both TAMP methods usually fail by greedily trying to grasp the keypoint. Without a very long horizon or the \signature{}, the planner often grasps with configurations that make reaching the goal impossible. The longer horizon used in $H=50$ helps alleviate this issue but is insufficient in many cases while also causing a 10x increase in planning time.

\subsection{Threading Environment} 

In the Threading environment, the objective is for the robot to thread the DOO through a series of fixtures in a specified order (e.g. ``fixture 1, then fixture 2, then fixture 3''), after which it should bring the keypoint to a goal region. The threading is described by a series of goal signatures $\sig_1,\dots,\sig_N$. This skill could be applied to installing cable harnesses in a car or electrical wiring in a building. One end of the DOO is fixed to the environment, and the robot is initially grasping some other location on the DOO. This environment is depicted in Figure \ref{fig:examples} B.

\begin{algorithm}[t]
\caption{DOO Threading with the \signature{}}
\label{alg:threading}
\begin{algorithmic}[1]
\Procedure{Threading}{$\state,\goalPoint,n_x,\sig_1,\dots,\sig_N$}
\State $j = 1$ \Comment{Threading subgoal index}
\For{$i < \maxIters$}
    \If{$j<N$} \Comment{threading subgoals}
        \State $\qvel = $MPPI$(\state,\sig_j, \goalCost)$
        \State $\state = f(\qvel)$ \Comment{Execute and get state}
        \If{{\color{green}disc penetrated}}
            \State $\locs^* = \planGrasp(\state, n_x,\graspCost, {\color{green}1}, \rev{\blist)}$
            \If{{\color{green} $\sig(\locs^*) == \sig_j$}}
                \State ExecuteGraspChange$(\locs^*)$
            \EndIf{}
        \EndIf{}
        \If{trapped} 
            \State $\locs^* = \planGrasp(\state, n_x, \graspCost, {\color{green}\loc-0.05}, \rev{\blist)}$
            \State ExecuteGraspChange$(\locs^*)$
        \EndIf{}
        \If{{\color{green} $\sig(\state) == \sig_j$}}
            \State $j = j + 1$ \Comment{next subgoal}
        \EndIf{}
    \Else{} \Comment{final point reaching}
        \State $\qvel = $MPPI$(\state,\goalPoint, \goalCost)$
        \State $\state = f(\qvel)$ \Comment{Execute and get state}
        \If{$\goalPoint - p(\kp) < \goalRadius$}
            \State break
        \EndIf{}
    \EndIf{}
\EndFor
\EndProcedure
\end{algorithmic}
\end{algorithm}

\rev{The method is detailed in Alg \ref{alg:threading}, and key differences to Alg \ref{alg:pointReaching} are highlighted in green. In MPC, the \signature{} is used in the same way as in Alg \ref{alg:pointReaching}, but the goal cost has been changed to match the new task. Additionally, we use the \signature{} to ensure certain goal signatures at each stage of threading. To reach a threading subgoal, we augment the goal cost (Eq \eqref{eq:pointReachingCost})} with the magnetic-field cost proposed in \cite{Weifu}. This uses the formula $\sum_{j=1}^{n_i}\Phi(\sk_i^j,\sk_i^{j'}, r)$ from Equation \eqref{eq:hsig} for the direction of the magnetic field, but where $r$ is the keypoint of the DOO. This causes the keypoint to follow virtual magnetic field lines through the fixture in the specified direction.

\rev{When} a threading subgoal is reached, and the planner returns a grasp which does not match $\goalSig$, we reject it and continue running MPPI to push the cable further through the fixture. This happens when there is no feasible grasp matching $\goalSig$ due to obstacles or reachability issues. Furthermore, we also check $\goalSig$ after executing the grasp to ensure that any deviations that occurred when executing the grasp plan do not change the \signature{}. To check when a threading subgoal is reached, we use the disc penetration check from \cite{Weifu}. The goal signatures $\sig_1,\dots,\sig_N$ are used in the grasp planning (3rd term in Eq \eqref{eq:graspCost}), but the blocklist is not \rev{(2nd term)}. Grasp sampling is restricted to alternating single-gripper grasps, which speeds up grasp planning. The keypoint location for grasp planning is also restricted \rev{to speed up planning}. It is chosen to be the tip ($\kp=1$) when a threading subgoal is reached, and further down the DOO than the current grasp when stuck ($\kp=\loc-0.05$). 

We compare our proposed method to the TAMP5 method described previously. In this environment, the TAMP method often chooses grasps that correctly thread through fixtures 1 and 2, because those grasps allow immediate progress towards the next subgoal. However, it often grasps incorrectly on fixture 3, which requires the robot to first reach further around and results in less immediate progress towards the next subgoal. We also adapted the method in \cite{Weifu} from a single floating gripper to our dual arm robot. As in our method, we use alternating single-gripper grasps. Instead of the more general trap detection method we use, this baseline checks the distance between the gripper and the fixture being threaded. This baseline fails similarly to the TAMP5 method, but additionally fails when MPPI is trapped but is outside the distance-to-fixture threshold. Success rates and trial times are shown in Table \ref{tab:threading}. Trials vary in the initial configuration of the robot, grasp location, DOO configuration, and in the positions of the fixtures. In the trials in which our method failed to complete the task, MPPI reached a joint configuration with one arm that prevented the other arm from grasping the DOO at or near the tip, as required by our method. This means the robot remained stuck until $\maxIters$ was reached.

\begin{table}
    \centering
    \begin{tabular}{cccc}
        Method & Success & Wall Time (m) & Sim Time (m) \\ \hline
        \signature{} & \textbf{42/50} & \textbf{8 (2)} & 1.3 (0.4) \\
        TAMP5 & 21/50 & 17 (3) & 1.3 (0.6) \\
        Wang et al. \cite{Weifu} & 12/50 & \textbf{8 (3)} & \textbf{1.0 (0.8)} \\
    \end{tabular}
    \caption{Results on the Threading task.}
    \label{tab:threading}
\end{table}


\subsection{Real World Threading}

We demonstrate a simplified version of the Threading task in the real world, as depicted in Figure \ref{fig:titleFig}. This shows the applicability of the proposed methods in the presence of significant calibration, perception, and dynamics modeling errors. We use CDCPD2 \cite{CDCPD2} to track the DOO and visual servoing from in-hand cameras to guide grasping. The environment geometry is specified manually, and the simulation dynamics were tuned to match the real world setup as closely as possible for the particular setup.

\section{DISCUSSION}

\begin{figure}
    \centering
    \includegraphics[width=\linewidth]{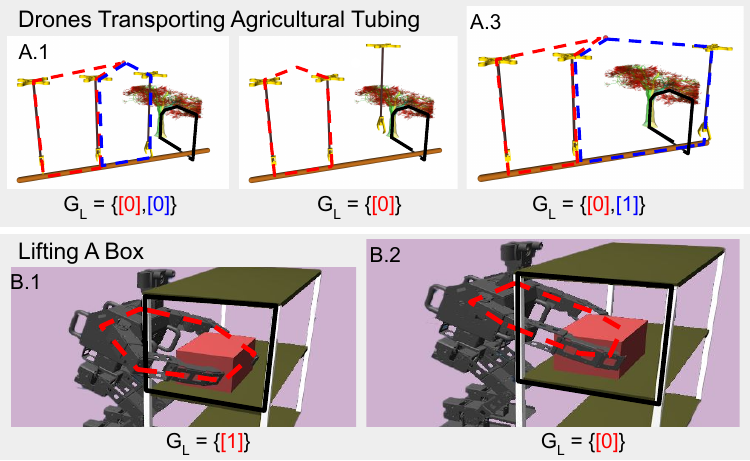}
    \caption{Additional examples where the \signature{} may be useful. (A) Drones lifting a large object can be treated similarly to a multi-armed robot. (B) Dual-arm grasping of large rigid objects can result in distinct signatures.}
    \label{fig:hypotheticalExamples}
\end{figure}

\rev{The planning methods described above are somewhat specific to their respective tasks, and the tasks represent a small subset of the domain of rope and cable manipulation. However, the \signature{} may be useful for other planning methods and other tasks, as shown in Figure \ref{fig:hypotheticalExamples}. In both of these examples, we can compute the \signature{} and see that it categorizes different states in ways that may be useful for planning.}

\rev{We use the \signature{} as part of the cost function in sampling-based MPC, but it could also be used as a constraint in RRT or A* planners (similar to \cite{Bhattacharya11}). Additionally, while the signature is composed of integers when all grasp and obstacle loops are closed, for incomplete loops they take on real values and have gradients which may be used in gradient-based planning methods. The \signature{} could also be used to constrain the topology of shape completion or scene reconstruction. For example, to ensure that when reconstructing the shape of a hook, it does not form a closed loop that would trap the robot arm or DOO.}

\section{CONCLUSION}

In this paper, we proposed the \signature{} which describes the topology of closed loops formed by grasping the DOO with respect to closed loops formed by stationary objects in the environment. Our \signature{} builds on the h-signature proposed in prior work on topological path planning. Furthermore, we describe an algorithm for manipulating DOOs that plans grasps based on the proposed \signature{}. In our experiments, we find that using the \signature{} improves task success and reduces planning times compared to a task and motion-planning method. Finally, we use the method to thread a cable and bring it to a goal region on a real robot.

\addtolength{\textheight}{0cm}

\bibliographystyle{unsrt}
\bibliography{references}

\end{document}